%% file: main.tex
\documentclass[10pt,twocolumn,letterpaper]{article}

\usepackage[pagenumbers]{cvpr} % To force page numbers, e.g. for an arXiv version

\usepackage{graphicx}
\usepackage{amsmath}
\usepackage{amssymb}
\usepackage{booktabs}

\usepackage{multirow}
\usepackage[dvipsnames]{xcolor}
\makeatletter
\@namedef{ver@everyshi.sty}{}
\makeatother
\usepackage{tikz}
\usepackage{pifont}
\usepackage{enumitem}

\usepackage[accsupp]{axessibility}  % Improves PDF readability for those with disabilities.

\newcommand{\tablestyle}[2]{\setlength{\tabcolsep}{#1}\renewcommand{\arraystretch}{#2}\centering}
\definecolor{green(pigment)}{rgb}{0.0, 0.65, 0.31}

\newcommand{\tb}[3]{\setlength{\tabcolsep}{#2mm}\begin{tabular}{#1}#3\end{tabular}}

\newcommand{\methodname}{CLIP-S$^4$}

\newcommand*\encircle[1]{\tikz[baseline=(char.base)]{
            \node[shape=circle,draw,inner sep=1pt] (char) {#1};}}
\newcommand{\xmark}{
\tikz[scale=0.16] {
  \draw[line width=0.7,line cap=round] (0,0) to [bend left=6] (1,1);
  \draw[line width=0.7,line cap=round] (0.2,0.95) to [bend right=3] (0.8,0.05);
}}

\newcommand*\rot{\rotatebox{70}}

\usepackage[pagebackref,breaklinks,colorlinks]{hyperref}

\usepackage[capitalize]{cleveref}
\crefname{section}{Sec.}{Secs.}
\Crefname{section}{Section}{Sections}
\Crefname{table}{Table}{Tables}
\crefname{table}{Tab.}{Tabs.}

\def\cvprPaperID{11808} % *** Enter the CVPR Paper ID here
\def\confName{CVPR}
\def\confYear{2023}

\begin{document}

\title{\methodname: Language-Guided \textit{S}elf-\textit{S}upervised \textit{S}emantic \textit{S}egmentation}

\author{
\tb{@{}cccc@{}}{8}{
Wenbin He&
Suphanut Jamonnak&
Liang Gou&
Liu Ren
}\\
Bosch Research North America \& Bosch Center for Artificial Intelligence (BCAI)\\
{\tt\small \{wenbin.he2, suphanut.jamonnak, liang.gou, liu.ren\}@us.bosch.com}
% \and
}
\maketitle

%%%%%%%%% ABSTRACT
\input{tex/0abstract}

%%%%%%%%% BODY TEXT
\input{tex/1introduction}
\input{tex/2related_work}
\input{tex/3method}
\input{tex/4experiments}
\input{tex/5conclusion}

%%%%%%%%% REFERENCES
{\small
\bibliographystyle{ieee_fullname}
\bibliography{main}
}

\end{document}

%% file: tex/0abstract.tex
\begin{abstract}

Existing semantic segmentation approaches are often limited by costly pixel-wise annotations and predefined classes.  In this work, we present {\methodname} that leverages self-supervised pixel representation learning and vision-language models to enable various semantic segmentation tasks (e.g., unsupervised, transfer learning, language-driven segmentation) without any human annotations and unknown class information. We first learn pixel embeddings with \textbf{pixel-segment contrastive learning} from different augmented views of images.  To further improve the pixel embeddings and enable language-driven semantic segmentation, we design two types of consistency guided by vision-language models: 1) \textbf{embedding consistency}, aligning our pixel embeddings to the joint feature space of a pre-trained vision-language model, CLIP~\cite{Radford2021}; and 2) \textbf{semantic consistency}, forcing our model to make the same predictions as CLIP over a set of carefully designed target classes with both known and unknown prototypes.  Thus, {\methodname} enables a new task of class-free semantic segmentation where no unknown class information is needed during training. As a result, our approach shows consistent and substantial performance improvement over four popular benchmarks compared with the state-of-the-art unsupervised and language-driven semantic segmentation methods.  More importantly, our method outperforms these methods on unknown class recognition by a large margin.

\end{abstract}

%% file: tex/1introduction.tex
\def\figIntro#1{
  \setlength{\abovecaptionskip}{6pt}
  \setlength{\belowcaptionskip}{-18pt}
  \begin{figure}[#1]
    \centering
    \includegraphics[width=\linewidth]{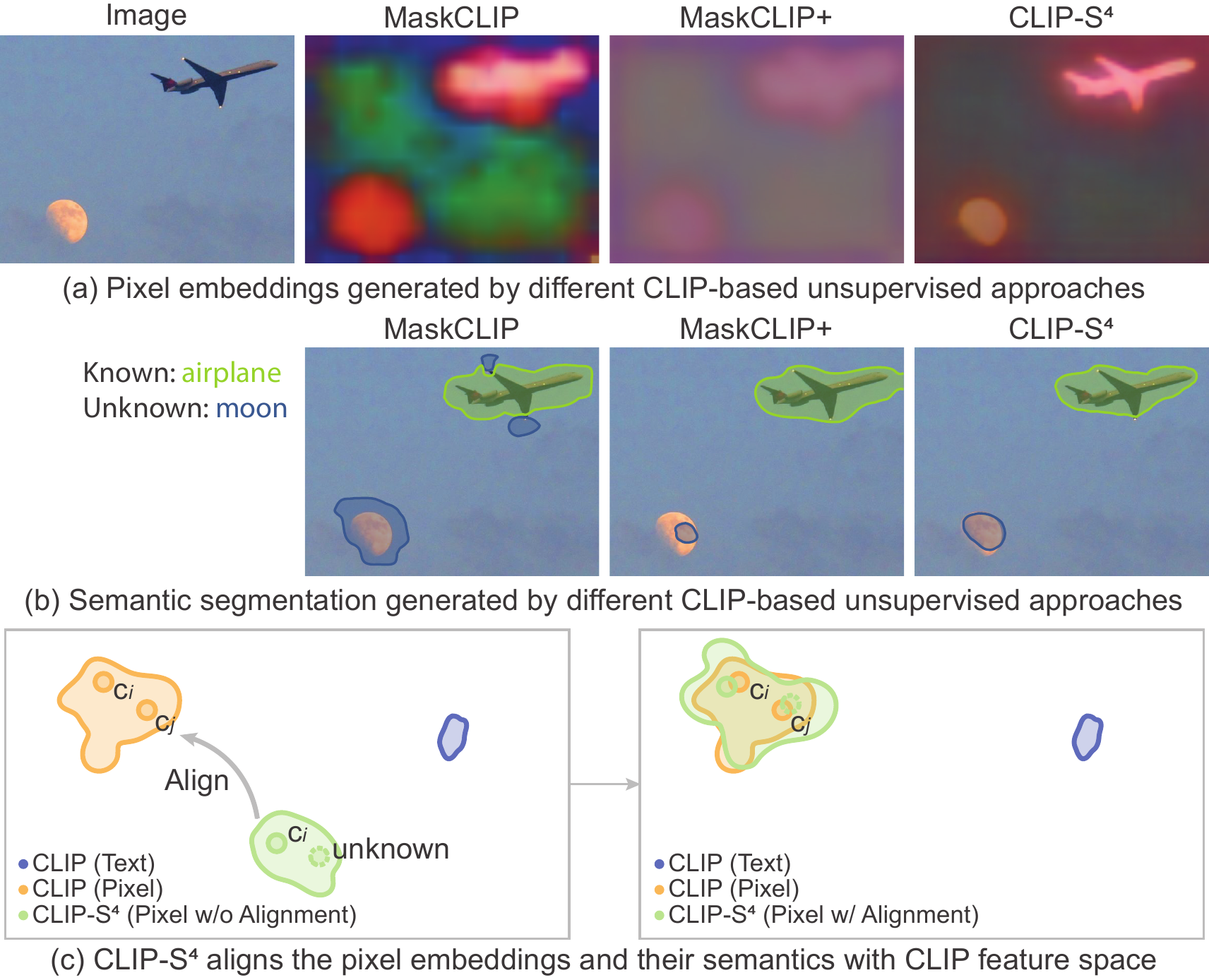}
    \caption{(a) Pixel embeddings from different CLIP-based unsupervised methods: Our method, {\methodname}, generates sharper and more coherent pixel embeddings than MaskCLIP~\cite{Zhou2022} and MaskCLIP+'s~\cite{Zhou2022}; (b) Language-driven semantic segmentation by different methods: {\methodname} can recognize challenging unknown classes (e.g., moon); (c) The key idea behind {\methodname}: aligning the pixel embeddings and their semantics with CLIP feature space.}
    \label{fig:intro}
  \end{figure}
  \setlength{\abovecaptionskip}{10pt}
  \setlength{\belowcaptionskip}{0pt}
}

\def\tabIntro#1{
\setlength{\abovecaptionskip}{6pt}
\setlength{\belowcaptionskip}{-16pt}
\begin{table}[#1]
  \centering
  \small
  \tablestyle{0.5pt}{0.95}
  \begin{tabular}{cccc|cc|c}
    \toprule
    & & \multicolumn{2}{c}{Known} & \multicolumn{2}{c}{Unknown} & \\
    & & \rot{Annot.} & \rot{Cls Name} & \rot{Annot.} & \rot{Cls Name} & \rot{Add. Info.} \\
    \midrule
    \multicolumn{2}{c}{Un/Self-supervised (~\cite{Hwang2019} etc.)} & \xmark & \xmark & \xmark & \xmark & Fine-Tuning \\
    \multicolumn{2}{c}{Supervised (~\cite{Long2015} etc.)} & $\checkmark$ & $\checkmark$ & N/A & N/A & N/A \\
    \multicolumn{2}{c}{Zero-shot (~\cite{Bucher2019} etc.)} & $\checkmark$ & $\checkmark$ & \xmark & $\checkmark$ & Word2Vec, etc. \\
    \midrule
    Language-\; & \textit{MaskCLIP+}~\cite{Zhou2022} & \xmark & $\checkmark$ & \xmark & $\checkmark$ & CLIP \\
    Driven & \textit{\methodname} & $\xmark$ & $\checkmark$ & \xmark & \xmark & CLIP \\
    \bottomrule
  \end{tabular}
  \caption{Comparison of information required for training over different tasks.  {\methodname} enables a new task called \textit{class-free semantic segmentation}.  Compared with MaskCLIP+~\cite{Zhou2022}, the new task assumes unknown class names are NOT given during training.}
  \label{tab:intro}
\end{table}
\setlength{\abovecaptionskip}{10pt}
\setlength{\belowcaptionskip}{0pt}
}

\section{Introduction}
\label{sec:introduction}

\figIntro{t}

Semantic segmentation aims to partition an input image into semantically meaningful regions and assign each region a semantic class label.  Recent advances in semantic segmentation~\cite{Long2015, Zhao2017, Chen2018} heavily rely on pixel-wise human annotations, which have two limitations.  First, acquiring pixel-wise annotations is extremely labor intensive and costly, which can take up to 1.5 hours to label one image~\cite{Papadopoulos2021}.  Second, human annotations are often limited to a set of predefined semantic classes, with which the learned models lack the ability to recognize unknown classes~\cite{Li2022}.

Various approaches have been proposed to tackle these limitations, among which we are inspired by two lines of recent research in particular.  First, for unsupervised semantic segmentation (i.e., without human annotations), self-supervised pixel representation learning approaches~\cite{Hwang2019, VanGansbeke2021, Ke2022, He2022, Hamilton2022} have shown promising results on popular unsupervised benchmarks.  The main idea is to extend self-supervised contrastive learning~\cite{Chen2020, He2020} from images to pixels by attracting each pixel's embedding to its positive pairs and repelling it from negative pairs. % The positive and negative pairs can be derived from different priors such as contours~\cite{Hwang2019, He2022}, hierarchical groups~\cite{Ke2022}, salience maps~\cite{VanGansbeke2021}, and pre-trained models~\cite{Hamilton2022}. 
The prior of pairs can be contours~\cite{Hwang2019, He2022}, hierarchical groups~\cite{Ke2022}, salience maps~\cite{VanGansbeke2021}, and pre-trained models~\cite{Hamilton2022}.
Although these approaches can group pixels into semantically meaningful clusters, human annotations are still needed to assign class labels to the clusters for semantic segmentation~\cite{Shin2022}.

Second, for unknown classes in semantic segmentation, large-scale vision-language models such as CLIP~\cite{Radford2021} have shown great potential.  %This line of research is called \textit{open-vocabulary segmentation}, which can segment images with arbitrary classes defined by texts.  
This line of research, called \textit{language-driven semantic segmentation}, aims to segment images with arbitrary classes defined by texts during testing time~\cite{Li2022, Shin2022, Xu2022, Zhou2022}. Among these methods, most still need training time annotations, such as pixel annotations ~\cite{Li2022} and captions~\cite{Xu2022}. Only a few recent work, MaskCLIP \& MaskCLIP+~\cite{Zhou2022} attempts to address this without using additional supervision: MaskCLIP directly extracts pixel embeddings correlated with texts from CLIP, but these pixel embeddings are coarse and noisy (\cref{fig:intro}a).  To address this issue, MaskCLIP+~\cite{Zhou2022} trains a segmentation model on the pseudo-labels generated by MaskCLIP for a set of predefined classes.  However, the pixel embeddings of MaskCLIP+ are distorted by the predefined classes (\cref{fig:intro}a), which limits its ability to recognize unknowns (\cref{fig:intro}b).  Also, it needs unknown class information during training, which hinders its real-world applications.

We propose a language-guided \underline{s}elf-\underline{s}upervised \underline{s}emantic \underline{s}egmentation approach, {\methodname}, which takes advantage of the strengths from both lines of research and addresses their limitations accordingly.  The key idea is to learn consistent pixel embeddings with respect to visual and conceptual semantics using self-supervised learning and the guidance of a vision-language model, CLIP.

Specifically, we first train pixel embeddings with \textit{pixel-segment contrastive learning} from different augmented image views~\cite{Hwang2019, Ke2022, He2022} such that images can be partitioned into visually meaningful regions.  To further improve pixel embedding quality and enable language-driven semantic segmentation, we introduce vision-language model guided consistency to regularize our model (\cref{fig:intro}c).  The consistency is enforced from two aspects: \textit{embedding consistency} and \textit{semantic consistency}.  First, embedding consistency aims to align the pixel embeddings generated by our model with the joint feature space of texts and images of CLIP by minimizing the distance between the pixel embeddings generated by our model and CLIP.  Second, semantic consistency forces our model to make the same prediction as CLIP over a set of carefully designed target
classes with both \textit{known} and \textit{unknown} prototypes. Note that unlike the previous methods~\cite{Li2022, Zhou2022} that use a predefined set of \textit{known classes}, {\methodname} also learns the representation of \textit{unknown classes} from images during training.

In the end, {\methodname} also enables a new task, namely \textit{class-free semantic segmentation}, as shown in~\cref{tab:intro}.  This new task does not need any human annotations and even assumes NO class names are given during training.  This is a more challenging task than the recent work~\cite{Zhou2022} that requires class names of both known and unknown.

\tabIntro{t}

In summary, the contributions of this paper are threefold:
\begin{itemize}
[topsep=0.2em]
\itemsep -0.3em
  \item We propose a self-supervised semantic segmentation approach that combines pixel-segment contrastive learning with the guidance of pre-trained vision language models.  Our method can generate high-quality pixel embeddings without any human annotations and be applied to a variety of semantic segmentation tasks.
  \item We open up new research potentials for language-driven semantic segmentation without any human annotations by introducing and addressing a new task of \textit{class-free semantic segmentation} (\cref{tab:intro}).  Unlike previous work that assumes all the class names are known during training, our method can discover unknown classes from unlabelled image data without even knowing unknown class names.
  \item Consistent and substantial gains are observed with our approach over the state-of-the-art unsupervised and language-driven semantic segmentation methods on four popular datasets.  More importantly, our method significantly outperforms the state-of-the-art on the segmentation of unknown classes.
\end{itemize}

%% file: tex/2related_work.tex
\def\figFramework#1{
  \setlength{\abovecaptionskip}{4pt}
  \setlength{\belowcaptionskip}{-16pt}
  \begin{figure*}[#1]
    \centering
    \includegraphics[width=\linewidth]{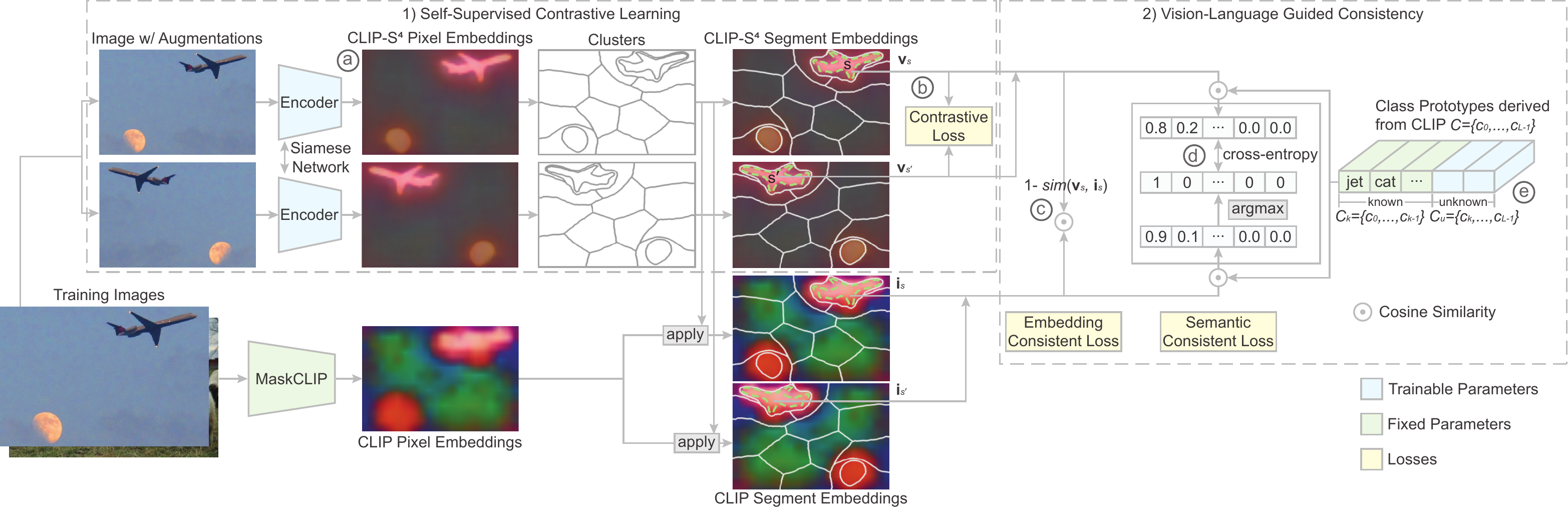}
    \caption{Framework of {\methodname}.  \protect\encircle{a} It starts with an encoder that maps images into pixel embeddings for semantic segmentation.  Then it follows with two components: 1) \textit{self-supervised contrastive learning} and 2) \textit{vision-language model guided consistency}.  Specifically, shown in \protect\encircle{b}, the self-supervised contrastive learning forces pixel embeddings to be consistent within visually coherent regions and among different augmented views of the same image.  For vision-language model guided consistency, this framework introduces \protect\encircle{c} \textit{embedding consistency} that aligns the pixel embeddings generated by our model with CLIP embeddings, and \protect\encircle{d} \textit{semantic consistency} that forces our model to make the same predictions as CLIP for a set of \textit{target classes} with both known and unknown prototypes.  For known classes, the prototypes are pre-computed and fixed during training.  For unknown classes, the prototypes are learned during training via clustering (\protect\encircle{e}).}
    \label{fig:framework}
  \end{figure*}
  \setlength{\abovecaptionskip}{10pt}
  \setlength{\belowcaptionskip}{0pt}
}

\figFramework{t}

\section{Related Work}
\label{sec:related_work}

\textbf{Unsupervised Semantic Segmentation.}  There are two groups of recent unsupervised semantic segmentation methods.  One group of methods learns to generate consistent pixel representations or predictions between different augmentation of images with the guidance of mutual information~\cite{Ji2019, Ouali2020},  clusters~\cite{Cho2021}, contours~\cite{Hwang2019, He2022}, hierarchical groups~\cite{Zhang2020, Ke2022}, and saliency masks~\cite{VanGansbeke2021}.  The other group of methods extracts dense features from pre-trained models based on saliency maps~\cite{Selvaraju2021}, augmentations~\cite{VanGansbeke2021a}, spectral decomposition~\cite{MelasKyriazi2022}, and feature correspondences~\cite{Hamilton2022}.  While these methods can generate pixel embeddings with semantically meaningful clusters, annotations are needed to assign class labels to the clusters (e.g., $k$-nearest neighbor search~\cite{Hwang2019} and Hungarian algorithm~\cite{VanGansbeke2021}).  Our work combines pixel-segment self-supervision with pre-trained vision-language models to enable semantic segmentation without any human annotations.

\textbf{Language-Driven Semantic Segmentation.}  Recently, vision-language models (e.g., CLIP~\cite{Radford2021}) trained on large-scale image-text datasets have shown great potential on various downstream tasks such as image synthesis~\cite{Jain2021, Vinker2022}, out-of-distribution detection~\cite{Esmaeilpour2022}, and object detection~\cite{Gu2022}.  To extend vision-language models for semantic segmentation, one active research, \textit{language-driven semantic segmentation}, aims to segment images with arbitrary unknown classes defined by texts during testing time~\cite{Li2022, Shin2022, Xu2022, Zhou2022}.  Some methods~\cite{Li2022, Wang2022} use pixel-wise annotations to train language-guided semantic segmentation models.  Other methods~\cite{Xu2022, Dong2022} perform large-scale pre-training on image-text pairs specifically for semantic segmentation.

By contrast, we directly use vision-language models that are pre-trained for classification tasks.  Along this line of research, a few approaches have been proposed~\cite{Zhou2022, Rao2022, Shin2022}.  The most relevant approach to our method is MaskCLIP~\cite{Zhou2022}, which extends the embeddings generated by pre-trained vision-language models from image to pixel level, but these embeddings are often coarse and noisy.  To address this issue, MaskCLIP+~\cite{Zhou2022} fine-tunes the pixel embeddings by the pseudo-labels of a specific set of classes on top of MaskCLIP. However, it needs unknown class names during training, which may not be possible in real-world cases.  Compared with~\cite{Zhou2022}, our method can recognize unknown classes without knowing any unknown class information during training time, and also learns fine-grained and sharper pixel embeddings with self-supervision.

%% file: tex/3method.tex
\def\figClassEmbeddings#1{
  \setlength{\abovecaptionskip}{6pt}
  \setlength{\belowcaptionskip}{-16pt}
  \begin{figure}[#1]
    \centering
    \includegraphics[width=\linewidth]{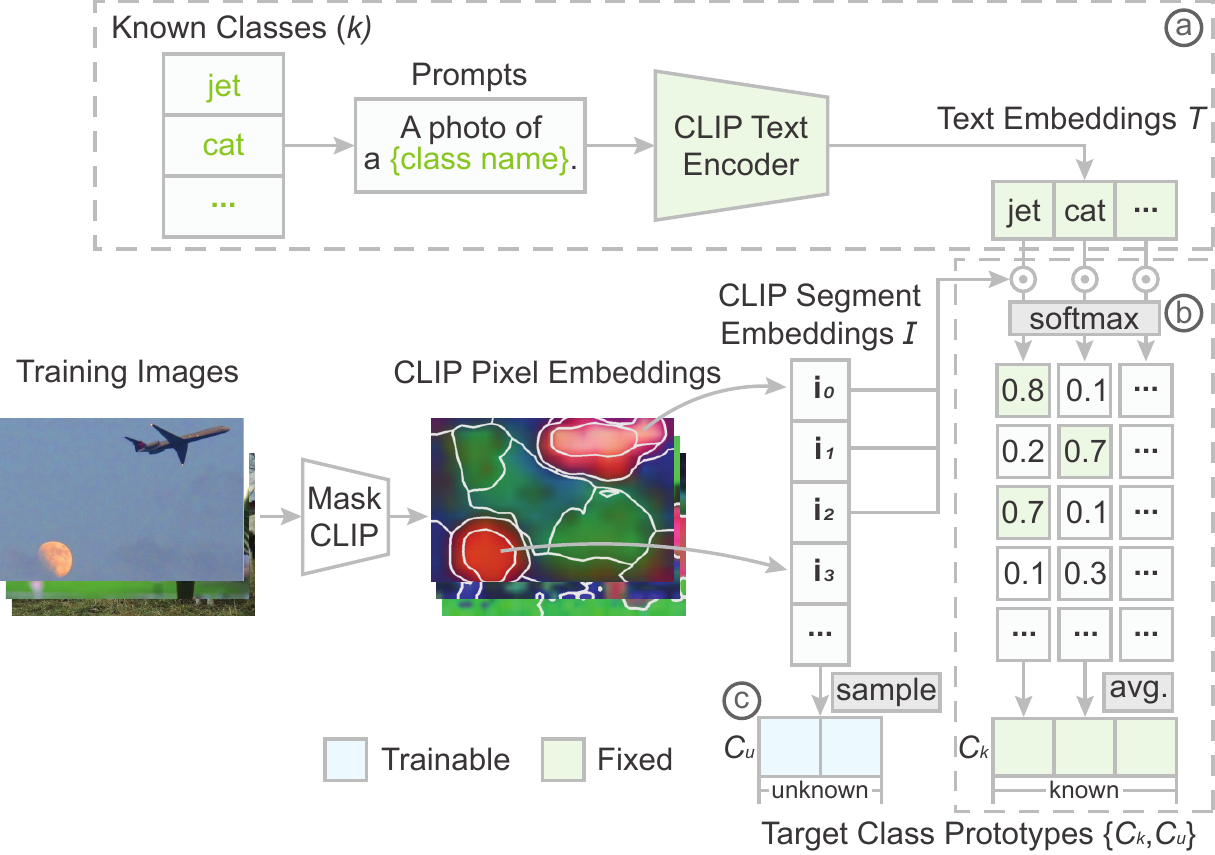}
    \caption{Computation of \textbf{target class prototypes} with both \textit{knowns} and \textit{unknowns}, $C=\{C_k, C_u\}$. For a set of known classes (e.g., bird, cat), we first obtain their CLIP text embeddings, $T$, via a set of prompt templates~\cite{Gu2022, Zhou2022} (shown in \protect\encircle{a}); then, we calculate the normalized (via softmax) similarity between text embeddings, $T$, and all segments' CLIP embeddings, $I$, from training images, and average the top-$m$ similar segments' CLIP embedding as the embedding prototype for this class, as shown in \protect\encircle{b}; For each unknown class, we randomly select the CLIP embedding of a segment as the initial prototype (shown in \protect\encircle{c}).}
    \label{fig:class_embeddings}
  \end{figure}
  \setlength{\abovecaptionskip}{10pt}
  \setlength{\belowcaptionskip}{0pt}
}

\section{Method}
\label{sec:method}

Our method (\cref{fig:framework}) segments images by learning a pixel embedding function with self-supervised contrastive learning and the guidance of a pre-trained vision-language model, CLIP.  We use self-supervised contrastive learning to force the consistency of pixel embeddings within visually coherent regions (e.g., superpixels) and among different augmented views of the same image (\cref{sec:contrastivelearning}).  We also introduce two vision-language model guided consistency (i.e., \textit{embedding consistency} and \textit{semantic consistency}) to further regularize the model (\cref{sec:languagelearning}).  The two components are complementary to each other.  On the one hand, contrastive learning mitigates the noise introduced by CLIP.  On the other hand, with the knowledge extracted from CLIP, the quality of the pixel embeddings can be improved.  More importantly, this approach enables us to perform language-driven semantic segmentation with our carefully designed \textit{target class prototypes} of both knowns and unknowns. In the following, we discuss the two components in detail.

\subsection{Pixel-Segment Contrastive Learning}
\label{sec:contrastivelearning}

We train a pixel embedding function to generate consistent pixel embeddings within visually coherent regions through pixel-segment contrastive learning~\cite{He2022, Ke2022}.  Specifically, the embedding function transforms each pixel $p$ of an image to a unit-length embedding vector $\mathbf{z}_p$ of dimension $d$ via a deep neural network.  The image is then partitioned into $\mathcal{|S|}$ segments by clustering the pixel embeddings.  The embedding $\mathbf{v}_s$ of each segment $s$ is calculated as the average of the pixel embeddings $\mathbf{v}_s=\sum_{p\in{s}}\mathbf{z}_p/|s|$, which is also normalized into a unit-length vector $\mathbf{v}_s=\mathbf{v}_s/\|\mathbf{v}_s\|$.  For each pixel $p$, the segments are grouped into two sets including a positive set $\mathcal{S^+}$ and a negative set $\mathcal{S^-}$.  The positive set $\mathcal{S^+}$ of a pixel contains segments within the same visually coherent region of the pixel.  Following the prior work~\cite{He2022, Ke2022}, the visually coherent region can be derived from super-pixels~\cite{Achanta2012} or contours~\cite{Arbelaez2011}.  We also use data augmentation (e.g., random cropping and color jittering) to generate consistent pixel embeddings between different augmented views of the same image.  Hence, segments within the same region of the pixel in any augmented views are considered as the positive set $\mathcal{S^+}$.  Other segments in the image and segments from other images in the same batch are included in the negative set $\mathcal{S^-}$.  The pixel embedding $\mathbf{z}_p$ is then attracted to the segments in positive set $\mathcal{S^+}$ and repelled from the segments in negative set $\mathcal{S^-}$ with \textit{contrastive loss}:
\begin{equation}
\mathcal{L}_t(p)=-log\frac{\sum_{s\in{\mathcal{S^+}}}{exp(sim(\mathbf{z}_p, \mathbf{v}_s)\kappa)}}{\sum_{s\in{\mathcal{S^+}\cup\mathcal{S^-}}}{exp(sim(\mathbf{z}_p, \mathbf{v}_s)\kappa)}},
\end{equation}
where $\kappa$ is the concentration constant and $sim(\mathbf{z}_p, \mathbf{v}_s)$ is the cosine similarity between the pixel embedding $\mathbf{z}_p$ and the segment embedding $\mathbf{v}_s$.

\subsection{Vision-Language Model Guided Consistency}
\label{sec:languagelearning}

To enable language-driven semantic segmentation and improve the quality of pixel embeddings, we use a pre-trained vision-language model such as CLIP~\cite{Radford2021} to guide the training of the pixel embedding function.  The key idea is to align the output space of our pixel embedding function consistent with the feature space of CLIP (\cref{fig:intro}c).  Specifically, two types of consistency are considered during training including \textit{embedding consistency} and \textit{semantic consistency}, which are detailed as follows.

\textbf{Embedding Consistency.}  Our goal is to align the pixel embeddings generated from our self-supervised method (the green contour in \cref{fig:intro}c) with CLIP's pixel embeddings (the orange contour in \cref{fig:intro}c).  This is done by minimizing the distance between the two pixel embedding spaces.

We first obtain the pixel embeddings of an input image from CLIP by modifying the attention-based pooling layer of the CLIP image encoder following~\cite{Zhou2022}.  Specifically, we 1) remove the query and key projection layers and 2) reformulate the value projection layer and the last linear layer as two consecutive fully connected layers.  In the following, we use ${clip\mbox{-}i(\cdot)}$ as the modified CLIP image encoder and ${clip\mbox{-}t(\cdot)}$ as CLIP text encoder.

Then we obtain the pixel embeddings of CLIP for different augmented views of the image.  Note that we use the original image to generate the CLIP pixel embeddings and perform augmentation afterwards to make sure that the CLIP pixel embeddings are consistent among different augmented views.  In the end, we minimize the distance of embeddings between \textbf{segments} instead of pixels from our self-supervised and CLIP embedding spaces.  This is because the pixel embeddings of CLIP are noisy (\cref{fig:framework}), which can be mitigated by aggregating over segments.  Hence, we use the pixel embeddings generated by our model to derive segments (clusters) and then apply them to the CLIP's pixel embeddings.  In the end, for each segment $s$, the \textit{embedding consistent loss} is defined as:
\begin{equation}
\mathcal{L}_e(s)=1-sim(\mathbf{v}_s, \mathbf{i}_s),
\end{equation}
where $\mathbf{v}_s$ and $\mathbf{i}_s$ are the segment embeddings derived from our embedding function and CLIP, respectively.  Here, $\mathbf{i}_s$ is the average of the CLIP pixel embedding from segment $s$, namely,  $\mathbf{i}_s=\sum_{p\in{s}}{clip\mbox{-}i(s)}/|s|$.

\figClassEmbeddings{t}

\textbf{Semantic Consistency}  In addition to embedding consistency, we introduce semantic consistency by forcing our model to make the same predictions of semantic classes as CLIP.  The rationale is that we can generate better pixel embeddings if they can form distinctive clusters corresponding to different semantic classes, as the goal of semantic segmentation is to perform pixel-wise classification.

Semantic consistency is achieved via a similar idea of pseudo-labeling~\cite{Sohn2020}.  Again, we force the semantic consistency at the segment level (not the pixel level) to reduce the noise in pseudo-labels.  Specifically, for each segment $s$, we first use CLIP to generate its pseudo-label $y_s$ over a set of target classes, which include both knowns and unknowns (we will introduce how to design these target classes later).  The pseudo-label is generated based on the highest similarity between the segment embedding $\mathbf{i}_s$ with a set of prototypes, $C=\{\mathbf{c}_l\}_0^{L-1}$, of the target classes in the pixel embedding space of CLIP, namely, $y_s=\mathbf{argmax}_{l\in{L}}(sim(\mathbf{i}_s, \mathbf{c}_l))$.

Then we define the \textit{semantic consistent loss} as the cross entropy between our model's prediction $\varphi(\mathbf{v}_s)$ over the target classes and the pseudo-label $y_s$:
\begin{equation}
\mathcal{L}_s(s)=\mathbf{H}(y_s, \varphi(\mathbf{v}_s)),
\end{equation}
where $\varphi(\mathbf{v}_s)=\mathbf{softmax}(sim(\mathbf{v}_s, C))$.

\textbf{Target Class Prototypes}  The design of \textit{target classes} and associated \textit{class prototypes}, $C=\{\mathbf{c}_l\}_0^{L-1}$, is crucial to achieve the semantic consistency.  Here, a class prototype, $\mathbf{c}_l$, is an embedding vector that can represent a class in an embedding space.  For example, it can be the mean vector of embeddings of all segments of a class ``car".  Currently, most existing methods~\cite{Li2022, Zhou2022} assume that the target classes are already predefined, which is not feasible in real-world use cases without any human annotations.  Thus, those methods cannot handle unknown classes hidden in the data.  To address this issue, we introduce two sets of class prototypes of \textit{known}, $C_k=\{\mathbf{c}_0, \dots, \mathbf{c}_{k-1}\}$, and \textit{unknown classes}, $C_u=\{\mathbf{c}_k, \dots, \mathbf{c}_{k+u}\}$, where the known classes are predefined by leveraging CLIP and the unknown classes are learned from image data during training.  Thus, we have $C=\{\mathbf{c}_l\}_0^{L-1} =\{\mathbf{c}_0, \dots, \mathbf{c}_{k-1}, \mathbf{c}_k, \dots, \mathbf{c}_{k+u}\}, L=k+u$.

For known classes, a natural choice is to use the text embeddings generated by CLIP as their class prototype embeddings~\cite{Li2022, Zhou2022}.  However, even though the text embeddings are trained to align with image/pixel embeddings~\cite{Radford2021}, there is still a huge gap between the text and image/pixel embeddings in the joint space of CLIP (\cref{fig:intro}c).  Therefore, it is challenging to learn meaningful unknown classes from image features when using text embeddings as class prototypes.  Hence, in this work, we use the prototype of CLIP pixel embeddings to represent each known class.

To this end, for a set of known classes (e.g., bird, cat), $K=\{0, \dots, k-1\}$, we first obtain their CLIP text embeddings, $T=\{\mathbf{t}_k\}=\{clip\mbox{-}t(k)\}$, via a set of prompt templates following~\cite{Gu2022, Zhou2022}, as shown in \cref{fig:class_embeddings}a.  We also get a set of CLIP segment embeddings, $I=\{\mathbf{i}_{\hat{s}}\}$, for all training images by a) feeding training images into the modified image encoder of CLIP to get pixel embeddings; b) clustering the pixel embeddings as segments, $\mathcal{\hat{S}}$; c) averaging the pixel embeddings in each segment, $\hat{s}$.  Hence, we have embeddings for each segment: $\mathbf{i}_{\hat{s}}=\sum_{p\in{\hat{s}}}{clip\mbox{-}i(p)}/|\hat{s}|$.  Then, we calculate the similarity between text embeddings of known classes, $T$, and all CLIP segment embeddings $I$, and normalize the similarities over all classes by softmax.  Finally, we average the top-$m$ similar segments' embedding as the embedding prototype for each class, %$C_k=\{\mathbf{c}_k\}=\{avg(top\mbox{-}m_{\hat{s}}\mathbf{softmax}(sim(\mathbf{t}_k, \mathbf{i}_{\hat{s}})))\}$
$C_k=\{\mathbf{c}_k\}=avg_m(top\mbox{-}m_{\hat{s}}(\mathbf{softmax}_k(sim(I, T))))$.

The prototype embeddings of the unknown classes, $C_u$, are randomly initialized by sampling the CLIP embeddings of all segments, namely ${C}_u=random({clip\mbox{-}i}(\mathcal{\hat{S}}), u)$, with a size of the unknown class of $u$ (\cref{fig:class_embeddings}c).  During training, the embedding $\mathbf{c}_u$ of each unknown class prototype is updated by minimizing its distance to all segments that are classified as this unknown class (similar to updating the centroids in $k$-means clustering):
\begin{equation}
\mathcal{L}_u=\sum_{s\in{S_u}}(1-sim(\mathbf{c}_u, {clip\mbox{-}i}(s)))/|S_u|,
\end{equation}
where $S_u$ are the segments classified as the unknown classes.  In this way, our model can also learn the pixel representation of unknown classes.

\subsection{Training and Inference}
\label{sec:training}

In summary, we train the pixel embedding function by combining the pixel-segment contrastive loss, embedding consistent loss, and semantic consistent loss:
\begin{equation}
\mathcal{L}=\mathcal{L}_t + \mathcal{L}_e + \mathcal{L}_s.
\end{equation}
During training, we also update the embeddings for the unknown classes with $\mathcal{L}_u$.

For inference, we use the trained model to generate pixel embeddings for each input image and use the pixel embeddings for different downstream tasks, including language-driven and unsupervised semantic segmentation.  For language-driven semantic segmentation, we first obtain the text embeddings of arbitrary inference classes by feeding the prompt-engineered texts into the text encoder of CLIP.  Then we assign each pixel with the class label whose text embedding is the closest to {\methodname} pixel embedding.  For unsupervised semantic segmentation, we follow previous work~\cite{Hwang2019, VanGansbeke2021} that uses $k$ nearest neighbor search or linear classifier to perform semantic segmentation.

%% file: tex/4experiments.tex
\def\tabLanguageDriven#1{
\setlength{\abovecaptionskip}{6pt}
\setlength{\belowcaptionskip}{-14pt}
\begin{table}[#1]
  \centering
  \small
  \tablestyle{3.8pt}{0.95}
  \begin{tabular}{lccc}
    \toprule
    \multirow{2}{*}{Method} & CLIP & Pascal & COCO- \\
    & Model & Context & Stuff \\
    \midrule
    & & mIoU & mIoU \\
    \midrule
    GroupViT~\cite{Xu2022} & - & 22.4 & - \\
    \midrule
    ReCo~\cite{Shin2022} & ResNet50x16 + & 26.6 & - \\
    ReCo+~\cite{Shin2022}$\dagger$ & ViT-L/14@336px & - & 18.4 \\
    \midrule
    \multirow{2}{*}{MaskCLIP~\cite{Zhou2022}} & ResNet50 & 18.6 & 10.6 \\
    & ViT-B/16 & 25.2 & 15.2 \\
    \midrule
    \multirow{2}{*}{MaskCLIP+~\cite{Zhou2022}$\dagger$} & ResNet50 & 23.4 & 13.9 \\
    & ViT-B/16 & 32.2 & 20.7 \\
    \midrule
    \multirow{2}{*}{{\methodname}$\dagger$} & ResNet50 & \textbf{28.5 \color{green(pigment)}(+5.1)} & \textbf{16.7 \color{green(pigment)}(+2.8)} \\
    & ViT-B/16 & \textbf{33.6 \color{green(pigment)}(+1.4)} & \textbf{22.1 \color{green(pigment)}(+1.4)} \\
    \bottomrule
  \end{tabular}
  \caption{\textbf{Language-guided semantic segmentation benchmarks (mIoU).} {\methodname} consistently outperforms the state-of-the-art methods on both Pascal Context and COCO-Stuff datasets with CLIP models of different backbones.  $\dagger$ indicates the models are fine-tuned on target datasets.}
  \label{tab:languagedriven}
\end{table}
\setlength{\abovecaptionskip}{10pt}
\setlength{\belowcaptionskip}{0pt}
}

\def\tabLanguageDrivenWithUnknown#1{
\setlength{\abovecaptionskip}{4pt}
\setlength{\belowcaptionskip}{-14pt}
\begin{table*}[#1]
  \centering
  \small
  \tablestyle{1pt}{0.95}
  \begin{tabular}{lcccccccccccc}
    \toprule
    \multirow{2}{*}{Method} & \multicolumn{3}{c}{fold0} & \multicolumn{3}{c}{fold1} & \multicolumn{3}{c}{fold2} & \multicolumn{3}{c}{fold3} \\
    & mIoU$_u$ & mIoU$_k$ & hIoU & mIoU$_u$ & mIoU$_k$ & hIoU & mIoU$_u$ & mIoU$_k$ & hIoU & mIoU$_u$ & mIoU$_k$ & hIoU \\
    % \midrule
    % \multicolumn{13}{l}{\textit{Pascal Context}} \\
    \midrule
    MaskCLIP & 29.7 & 23.7 & 26.3 & \textbf{23.7} & 25.7 & 24.6 & \textbf{23.9} & 25.7 & 24.7 & 23.4 & 25.8 & 24.5 \\
    MaskCLIP+ & 3.6 & 28.5 & 6.3 & 3.0 & 29.2 & 5.4 & 4.8 & 29.2 & 8.2 & 4.5 & 29.9 & 7.8 \\
    \methodname & \textbf{32.0$\pm$0.8} & \textbf{29.4$\pm$0.3} & \textbf{30.6$\pm$0.5} & 22.3$\pm$0.9 & \textbf{32.8$\pm$0.4} & \textbf{26.5$\pm$0.6} & 22.4$\pm$0.5 & \textbf{32.1$\pm$0.5} & \textbf{26.4$\pm$0.4} & \textbf{28.6$\pm$0.8} & \textbf{31.5$\pm$0.2} & \textbf{30.0$\pm$0.5} \\
    \textit{vs. MaskCLIP+} & \textbf{\color{green(pigment)}+28.4} & \textbf{\color{green(pigment)}+0.9} & \textbf{\color{green(pigment)}+24.3} & \textbf{\color{green(pigment)}+19.3} & \textbf{\color{green(pigment)}+3.6} & \textbf{\color{green(pigment)}+21.1} & \textbf{\color{green(pigment)}+17.6} & \textbf{\color{green(pigment)}+2.9} &
    \textbf{\color{green(pigment)}+18.2} & \textbf{\color{green(pigment)}+24.1} & \textbf{\color{green(pigment)}+1.6} & \textbf{\color{green(pigment)}+22.2} \\
    % \midrule
    % \multicolumn{13}{l}{\textit{COCO-Stuff}} \\
    % \midrule
    % MaskCLIP & & & & & & & & & & & & \\
    % MaskCLIP+ & & & & & & & & & & & & \\
    % \methodname & & & & & & & & & & & & \\
    % \textit{vs. baseline} & & & & & & & & & & & & \\
    \bottomrule
  \end{tabular}
  \caption{\textbf{Language-guided semantic segmentation benchmarks (mIoU) for unknown classes.}  The classes of Pascal Context are split into 4 folds with around 15 classes each fold.  For each experiment, classes of one fold are considered as unknown.  The performance of {\methodname} is averaged over 5 runs with randomly initialized unknown class embeddings.  {\methodname} significantly outperforms MaskCLIP+ on unknown classes.  Meanwhile, {\methodname} archives consistent gains on known classes over MaskCLIP and MaskCLIP+, and hence leads to better overall performance.}
  \label{tab:languagedrivenwithunknown}
\end{table*}
\setlength{\abovecaptionskip}{10pt}
\setlength{\belowcaptionskip}{0pt}
}

\def\tabUnsupervised#1{
\setlength{\abovecaptionskip}{6pt}
\setlength{\belowcaptionskip}{-18pt}
\begin{table}[#1]
  \centering
  \small
  \tablestyle{6pt}{0.95}
  \begin{tabular}{lcc}
    \toprule
    \multirow{2}{*}{Method} & mIoU & mIoU \\
    & $k$-NN & Linear Classifier \\
    % \midrule
    % Inst. Discr. & - & 26.8 \\
    % MoCo v2 & - & 45.0 \\
    % InfoMin & - & 45.2 \\
    % SWAV & - & 50.7 \\
    \midrule
    IIC~\cite{Ji2019} & - & 28.0 \\
    SegSort~\cite{Hwang2019} & 47.3 & 55.4 \\
    Hierarch. Group.~\cite{Zhang2020}  & - & 48.8 \\
    MaskContrast (Sup.)~\cite{VanGansbeke2021} & 53.9 & 63.9 \\
    ConceptContrast~\cite{He2022} & 58.8 & 60.4 \\
    HSG~\cite{Ke2022} & 61.7 & - \\
    \midrule
    MaskCLIP~\cite{Zhou2022} & 67.3 & 69.5 \\
    MaskCLIP+~\cite{Zhou2022} & 65.1 & 70.0 \\
    \midrule
    {\methodname} & \textbf{72.0 \color{green(pigment)}(+4.7)} & \textbf{73.0 \color{green(pigment)}(+3.0)} \\
    \bottomrule
  \end{tabular}
  \caption{\textbf{Unsupervised semantic segmentation benchmarks (mIoU) on Pascal VOC 2012.} {\methodname} consistently outperforms the state-of-the-art methods on both $k$-NN search and linear classification.}
  \label{tab:unsupervised}
\end{table}
\setlength{\abovecaptionskip}{10pt}
\setlength{\belowcaptionskip}{0pt}
}

\def\tabDavis#1{
\setlength{\abovecaptionskip}{6pt}
\setlength{\belowcaptionskip}{-2pt}
\begin{table}[#1]
  \centering
  \small
  \tablestyle{8pt}{0.95}
  \begin{tabular}{lcc}
    \toprule
    Method & $\mathcal{J}$(Mean)$\uparrow$ & $\mathcal{F}$(Mean)$\uparrow$ \\
    % \midrule
    % MaskTrack (fine-tuned)~\cite{Perazzi2017} & 51.2 & 57.3 \\
    % OSVOS (fine-tuned)~\cite{Caelles2017} & 55.1 & 62.1 \\
    \midrule
    MaskTrack-B~\cite{Perazzi2017} & 35.3 & 36.4 \\
    OSVOS-B~\cite{Caelles2017} & 18.5 & 30.0 \\
    \midrule
    Video Colorization~\cite{Vondrick2018} & 34.6 & 32.7 \\
    CycleTime~\cite{Wang2019} & 41.9 & 39.4 \\
    mgPFF~\cite{Kong2019} & 42.2 & 46.9 \\
    Hierarch. Group.~\cite{Zhang2020} & 47.1 & 48.9 \\
    MaskContrast (Sup.)~\cite{VanGansbeke2021} & 34.3 & 36.7 \\
    ConceptContrast~\cite{He2022} & 50.4 & 53.9 \\
    \midrule
    MaskCLIP~\cite{Zhou2022} & 48.1 & 49.2 \\
    MaskCLIP+~\cite{Zhou2022} & 42.6 & 44.2 \\
    \midrule
    \methodname & \textbf{52.3 \color{green(pigment)}(+1.9)} & \textbf{56.8 \color{green(pigment)}(+2.9)} \\
    \bottomrule
  \end{tabular}
  \caption{\textbf{Quantitative evaluation of instance mask tracking on the DAVIS-2017 validation set.}  The performance is measured by the region similarity $\mathcal{J}$ (IoU) and the contour-based accuracy $\mathcal{F}$ defined by~\cite{Perazzi2016}.  Our method outperforms existing supervised, unsupervised, and language-guided approaches on both metrics.}
  \label{tab:davis}
\end{table}
\setlength{\abovecaptionskip}{10pt}
\setlength{\belowcaptionskip}{0pt}
}

\def\tabLosses#1{
\setlength{\abovecaptionskip}{6pt}
\setlength{\belowcaptionskip}{-14pt}
\begin{table}[#1]
  \centering
  \small
  \tablestyle{9pt}{0.95}
  \begin{tabular}{cccccc}
    \toprule
    $\mathcal{L}_t$ & $\mathcal{L}_e$ & $\mathcal{L}_s$ & pAcc & mIoU & $avgsim$ \\
    \midrule
    $\checkmark$ & - & - & 1.6 & 0.5 & -0.01 \\
    $\checkmark$ & $\checkmark$ & - & 48.1 & 24.3 & 0.79 \\
    $\checkmark$ & - & $\checkmark$ & 52.3 & 32.9 & 0.33 \\
    - & $\checkmark$ & $\checkmark$ & 48.6 & 31.3 & - \\
    $\checkmark$ & $\checkmark$ & $\checkmark$ & \textbf{53.7} & \textbf{33.6} & 0.66 \\
    \bottomrule
  \end{tabular}
  \caption{\textbf{Ablation study on the contribution of each loss of {\methodname}}.  Experimented on language-guided semantic segmentation of Pascal Context.  $avgsim$ represents the average cosine similarity between segment embeddings generated by \methodname and CLIP.  By combining the embedding and semantic consistent losses $\mathcal{L}_e$ and $\mathcal{L}_s$, {\methodname} archives better semantic segmentation performance while maintaining the alignment with CLIP's embeddings.}
  \label{tab:losses}
\end{table}
\setlength{\abovecaptionskip}{10pt}
\setlength{\belowcaptionskip}{0pt}
}

\def\tabParameters#1{
\setlength{\abovecaptionskip}{6pt}
\setlength{\belowcaptionskip}{-18pt}
\begin{table}[#1]
\begin{minipage}[t][][b]{.22\textwidth}
  \tablestyle{4pt}{0.95}
  \begin{tabular}{lc}
    \toprule
    \#unknowns($u$) & mIoU \\
    \midrule
    16 & 71.6 \\
    32 & 71.9 \\
    64 & 72.0 \\
    128 & 72.2 \\
    256 & 71.9 \\
    \bottomrule
  \end{tabular}
  \caption{\textbf{Ablation study} on the influence of the number of unknown class prototypes.}
  \label{tab:nunknown}
\end{minipage}\hspace{3pt}
\begin{minipage}[t][][b]{.24\textwidth}
  \tablestyle{4pt}{0.95}
  \begin{tabular}{lc}
    \toprule
    top-$m$ segments  & $avgsim$ \\
    \midrule
    1 & 0.921 \\
    4 & 0.977 \\
    16 & 0.996 \\
    64 & 0.997 \\
    256 & 0.993 \\
    \bottomrule
  \end{tabular}
  \caption{\textbf{Ablation study} on different numbers of top-$m$ segments for class prototypes.}
  \label{tab:nsamples}
\end{minipage}
\end{table}
\setlength{\abovecaptionskip}{10pt}
\setlength{\belowcaptionskip}{0pt}
}

\def\figCompare#1{
  \setlength{\abovecaptionskip}{6pt}
  \setlength{\belowcaptionskip}{-18pt}
  \begin{figure}[#1]
    \centering
    \includegraphics[width=\linewidth]{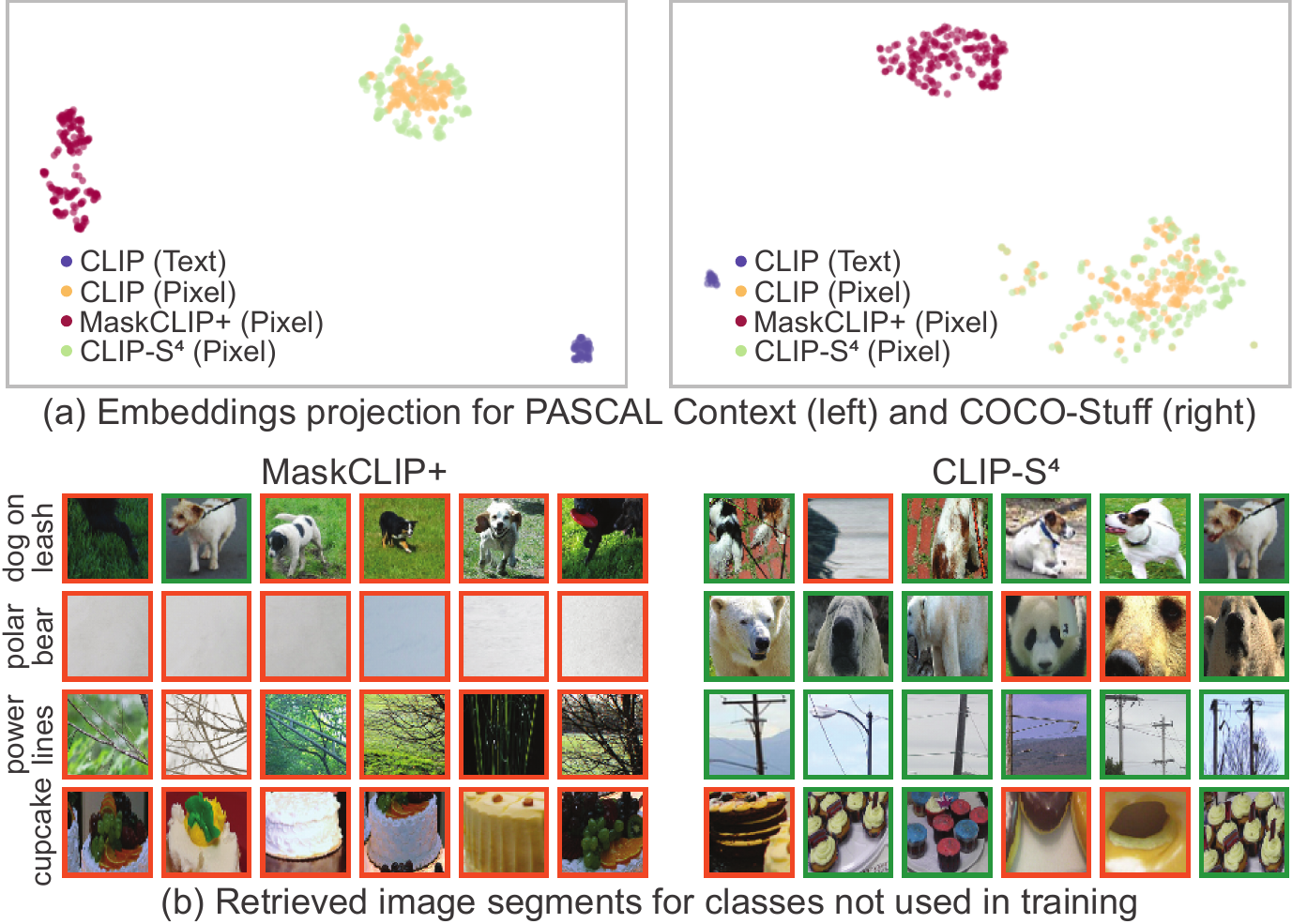}
    \caption{(a) Projection of pixel embeddings generated by CLIP, MaskCLIP+, and {\methodname} trained on Pascal Context with CLIP's ViT-B/16 model (left) and COCO-Stuff with CLIP's ResNet50 model (right).  MaskCLIP+ distorts pixel embeddings with respect to the given text embeddings, while {\methodname} aligns pixel embeddings with the pre-trained CLIP model.  (b) Image segments retrieved for classes that are not used during training.  Correct and incorrect retrievals are outlined in green and orange, respectively.  Compared with MaskCLIP+, {\methodname} retrieves images that are more closely related to the classes.}
    \label{fig:compare}
  \end{figure}
  \setlength{\abovecaptionskip}{10pt}
  \setlength{\belowcaptionskip}{0pt}
}

\section{Experiments}
\label{sec:experiments}

% \subsection{Experimental Setup}
% \label{sec:experimental_setup}

We evaluate our model on three tasks: 1) language-driven semantic segmentation for both known and unknown classes; 2) unsupervised semantic segmentation with $k$-means clustering/linear classification; 3) transfer learning of generated pixel embeddings for instance mask tracking.  We also conduct ablation studies to understand the components of our model.

\subsection{Datasets}
\label{sec:datasets}

\textbf{Pascal VOC 2012}~\cite{Everingham2010} contains 20 object classes and a background class.  It has 1,464 and 1,449 images for training and validation, respectively.  Following common practice~\cite{Long2015, Zhao2017}, we augment the training data with additional annotations~\cite{Hariharan2011}, resulting in 10,582 training images.

\textbf{Pascal Context}~\cite{Mottaghi2014} extends Pascal VOC 2010~\cite{Everingham2010} with additional annotations on 4,998 training and 5,105 validation images.  Following the prior work~\cite{Zhou2022}, we use the most common 59 classes for evaluation.

\textbf{COCO-Stuff}~\cite{Caesar2018} labels MS COCO~\cite{Lin2014} with 171 object/stuff classes.  It contains 118,287 and 5,000 images for training and validation, respectively.

\textbf{DAVIS 2017}~\cite{Perazzi2016} contains video sequences for instance mask tracking.  Following the prior work~\cite{Zhang2020, He2022}, we train pixel embeddings on Pascal VOC 2012 and evaluate the validation sequences without fine-tuning.

It is worth mentioning that \textbf{no ground truth} labels of any datasets are used during training.  Instead, we perform self-supervised learning on pseudo segments generated by contour detectors and owt-ucm~\cite{Arbelaez2011}.  Two contour detectors are used including HED~\cite{Xie2015} for Pascal VOC 2012 and Pascal Context and PMI~\cite{Isola2014} for COCO-Stuff.

\subsection{Implementation Details}
\label{sec:implementation_details}

For self-supervised contrastive learning, images are augmented with the same set of data augmentations as SimCLR~\cite{Chen2020}, including random resizing, cropping, flipping, color jittering, and Gaussian blurring.  The concentration constant $\kappa$ is set to 10, and the number of segments is set to 36 for each augmented view.

For vision-language guidance, we use pre-trained CLIP models~\cite{Radford2021} with modified image encoders following~\cite{Zhou2022}.  We use prompt-engineered texts with 85 prompt templates to generate text embeddings following~\cite{Gu2022, Zhou2022}.  We use the average embedding of the top 32 segments of high probabilities as the prototype of each known class.  We set the number of unknown classes to $u=64$.

Following the prior work~\cite{Hwang2019, He2022, Ke2022}, we use PSPNet~\cite{Zhao2017} with a dilated ResNet-50~\cite{He2016} backbone as the network architecture.  The backbone is pre-trained on the ImageNet~\cite{Deng2009} dataset.  We train our model on Pascal VOC 2012 and Pascal Context for 20k iterations and on COCO-Stuff for 80k iterations.  We set the batch size to 8 with additional memory banks that cache the segment embeddings of the previous 2 batches.  We set the initial learning rate to 0.001 and decay it with a polynomial learning rate policy.  We use the CLIP model trained with ViT-B/16 backbone unless otherwise stated.

\subsection{Language-Driven Semantic Segmentation}
\label{sec:language_driven}

For language-driven semantic segmentation, no human annotations are used for either training or inference.  At the inference time, each pixel is assigned an arbitrarily given class label whose CLIP text embedding is the closest to this pixel's {\methodname} embedding.

\tabLanguageDriven{t}

We first compare the performance of our method with the state-of-the-art language-driven semantic segmentation approaches~\cite{Shin2022, Xu2022, Zhou2022} on the Pascal Context and COCO-Stuff datasets.  The performance is evaluated with the mean Intersection over Union (mIoU).  For MaskCLIP and MaskCLIP+~\cite{Zhou2022}, we obtain the results using the same hyper-parameter setting as our approach with CLIP models of two different backbones of ResNet50 and ViT-B/16.  Meanwhile, GroupViT~\cite{Xu2022}, ReCo~\cite{Shin2022}, and ReCo+~\cite{Shin2022} use completely different training mechanisms compared with our method.  GroupViT is trained on image-caption pairs, and ReCo/ReCo+ combines image retrieval and co-segmentation.  For comparison, we take the best results from~\cite{Xu2022, Shin2022} for GroupViT, ReCo, and ReCo+.  Tabel~\ref{tab:languagedriven} shows the benchmarking results of the aforementioned methods.  Our method consistently outperforms the state-of-the-art on both datasets with CLIP models of different backbones.

\tabLanguageDrivenWithUnknown{t}

To evaluate models' performance for \textit{class-free semantic segmentation} with both known and unknown classes,  we split the 59 classes of Pascal Context into 4 folds, where each fold includes around 15 classes.  For each experiment, classes from one fold are considered as unknown and \textbf{excluded during training}.  The mIoUs of known and unknown classes, as well as their harmonic mean (hIoU) are reported in~\cref{tab:languagedrivenwithunknown}.  The performance of our method is averaged across 5 runs with randomly initialized prototypes of unknown classes.  Our method achieves significant gains over MaskCLIP+ on unknown classes, which are comparable to MaskCLIP.   Also, our method outperforms both MaskCLIP and MaskCLIP+ on known classes, which leads to better overall performance.

Qualitatively, the visualization in \cref{fig:compare}a offers us some insights into why our approach can achieve better results: our model yields \textit{consistent} embeddings aligned with the pre-trained CLIP model.  \cref{fig:compare}a visualizes the projection of segment embeddings generated by different methods on Pascal Context and COCO-Stuff.  We observe that segment embeddings generated by MaskCLIP+ are distorted by the given text embeddings.  Meanwhile, {\methodname} generates segment embeddings that are well aligned with the segment embeddings derived from the pre-trained CLIP model.  Hence, segment embeddings generated by {\methodname} can better capture both known and unknown classes.  \cref{fig:compare}b shows image segments retrieved from the COCO-Stuff validation set using MaskCLIP+ and {\methodname} for a set of classes that are not used in training.  For each class, we obtain its text embedding and compare it with segment embeddings to obtain top retrievals from both methods.  Due to better alignment with the pre-trained CLIP model, {\methodname} retrieves images that are more closely related to the unknown classes compared with MaskCLIP+.

\figCompare{t}

\subsection{Unsupervised Semantic Segmentation}
\label{sec:unsupervised}

\tabUnsupervised{t}

To study whether {\methodname} can generate pixel embeddings that form distinctive clusters, we evaluate {\methodname} on the unsupervised semantic segmentation task for Pascal VOC 2012.  To derive semantic segmentation from pixel embeddings, we use and test two approaches, including $k$ nearest neighbor ($k$-NN) search~\cite{Hwang2019} and linear classification~\cite{VanGansbeke2021}.  For $k$-NN search, we assign each segment a class label by the majority vote of its nearest neighbors from the training set following~\cite{Hwang2019}.  For linear classification, we train a linear classifier on the learned pixel embeddings following~\cite{VanGansbeke2021}.

We compare our method with the state-of-the-art unsupervised and language-guided semantic segmentation approaches.  We train the state-of-the-art models using the same hyper-parameter setting as our approach except for IIC~\cite{Ji2019} and Hierarchical Grouping~\cite{Zhang2020} as they use different training mechanisms.  For comparison, we take the best results for IIC and Hierarchical Grouping.  The benchmark results are shown in~\cref{tab:unsupervised}.  With the vision-language guidance, our method achieves significant gains compared with the previous non-CLIP-based approaches (i.e., +9\% for both $k$-NN search and linear classification).  Meanwhile, our method also outperforms the language-guided semantic segmentation approaches by a large margin.

\subsection{Instance Mask Tracking}
\label{sec:instance}

\tabDavis{t}

We evaluate the transferability of pixel embeddings learned from the Pascal VOC 2012 dataset.  We use the pixel embeddings to track instance masks in the DAVIS 2017 validation set, where the instance masks at the first frame are given for each video.  Following the prior work~\cite{Zhao2017}, we use the similarity between pixel embeddings cross frames to propagate the instance masks to the rest of the video frames.  We evaluate the performance using the region similarity $\mathcal{J}$ (IoU) and the contour-based accuracy $\mathcal{F}$ defined by~\cite{Perazzi2016}.

\tabLosses{t}

We compare our method with existing supervised~\cite{Perazzi2017, Caelles2017}, unsupervised~\cite{Vondrick2018, Wang2019, Kong2019, Zhang2020, VanGansbeke2021, He2022}, and language-guided~\cite{Zhou2022} approaches (\cref{tab:davis}).  Though not trained on any video sequences, our method outperforms the existing approaches by more than 1.9\% and 2.9\% in terms of the region similarity $\mathcal{J}$ and contour accuracy $\mathcal{F}$, respectively.  Note that the pixel embeddings generated by MaskCLIP+~\cite{Zhou2022} are distorted by the classes from Pascal VOC 2012, which hinder their transferability.

\subsection{Ablation Study}
\label{sec:ablation_study}

We study the contribution of different losses of our method using the Pascal Context dataset and the language-guided semantic segmentation task.  The performance is evaluated with pixel accuracy (pAcc) and mIoU.  We also calculate the average cosine similarity ($avgsim$) between our segment embeddings and CLIP's segment embeddings to quantify the alignment.  \cref{tab:losses} shows the study results.  We observe that by introducing embedding consistent loss $\mathcal{L}_e$ the learned segment embeddings are well aligned with CLIP's embeddings with an average cosine similarity of 0.79.  However, the learned segment embeddings do not perform well on the language-guided semantic segmentation task (24.3 vs. 33.6), because the segment embeddings are not optimized to classify target classes.  Meanwhile, by using semantic consistent loss $\mathcal{L}_s$ without embedding consistent loss, the learned segment embeddings have the discriminative power to classify different classes but are not aligned with CLIP's embeddings as the average cosine similarity is 0.33.  As a result, the segment embeddings are limited to the target classes used during training.  Hence, we combine $\mathcal{L}_e$ and $\mathcal{L}_s$ to balance the discriminative power over target classes and the alignment with CLIP.  Meanwhile, we observe that with pixel-segment contrastive learning, the model can archive better performance.

\tabParameters{t}

Also, we study the influence of the number of unknown class prototypes on the Pascal VOC dataset for the unsupervised semantic segmentation task.  The results~\cref{tab:nunknown} show that the semantic segmentation performance is robust to the tested number of unknown class prototypes as the mIoU varies only 0.6\%.  Furthermore, we investigate how the size of top-$m$ segments impacts the embeddings of class prototypes.  We compare the embeddings of class prototypes generated with different numbers of top-$m$ segments on the Pascal VOC dataset.  We use the embeddings of class prototypes generated with $m=32$ segments as the reference, and compute the cosine similarity between the reference prototypes embeddings and ones with different top-$m$ segments.  For each case, the cosine similarity is averaged over all class prototypes.  We observe that the embeddings of class prototypes are relatively stable if moderate top-$m$ segments (e.g., $m=32$ in this work) are used (\cref{tab:nsamples}).

%% file: tex/5conclusion.tex
\section{Conclusion}
\label{sec:conclusion}

We propose \methodname, a novel pixel representation learning approach for semantic segmentation.  Our method combines self-supervised contrastive learning and guidance of CLIP to learn consistent pixel embeddings with respect to visual and conceptual semantics.  Our experiments on popular semantic segmentation benchmarks demonstrate consistent gains over the state-of-the-art unsupervised semantic segmentation and language-driven semantic segmentation methods, especially for unknown classes.